\documentclass[times,twocolumn,final,5p]{elsarticle}

\usepackage{framed,multirow}

\usepackage{amssymb}
\usepackage{latexsym}

\usepackage{url}
\usepackage{xcolor}
\usepackage{times}
\usepackage{soul}

\usepackage{hyperref}

\usepackage{amsfonts}
\usepackage{subfigure}

\usepackage{amssymb,bbding}

\usepackage{graphicx}
\usepackage{amsmath}
\usepackage{amsthm}
\usepackage{algorithm}
\usepackage{algorithmic}
\usepackage[switch]{lineno}

\usepackage{tabularx}
\usepackage{booktabs}
\usepackage{array}
\usepackage{siunitx}
\usepackage{wasysym}

\usepackage{float}

\definecolor{newcolor}{rgb}{.8,.349,.1}

\begin{document}
	
	
	\begin{frontmatter}
		
		\title{MonoPCC: Photometric-invariant Cycle Constraint for Monocular Depth Estimation of Endoscopic Images}%
		
		\author[1]{Zhiwei Wang\fnref{fn1}}
		
		\author[1]{Ying Zhou\fnref{fn1}}
		
		\author[1]{Shiquan He}
		
		\author[2]{Ting Li}
		\author[2]{Fan Huang}
		
		\author[3]{Qiang Ding}
		\author[3]{Xinxia Feng}
		\author[3]{Mei Liu}
		
		\author[1]{Qiang Li\corref{cor1}}
		\ead{liqiang8@hust.edu.cn}
		
		\cortext[cor1]{Corresponding author.}
		\fntext[fn1]{Equal contribution.}

		\address[1]{Britton Chance Center for Biomedical Photonics, Wuhan National Laboratory for Optoelectronics, Huazhong University of Science and Technology, Wuhan, 430074, China.}
		\address[2]{Wuhan United Imaging Healthcare Surgical Technology Co., Ltd., Wuhan, 430074, China.}
		\address[3]{Department of Gastroenterology, Tongji Hospital, Tongji Medical College, Huazhong University of Science and Technology, Wuhan, 430074, China}
		
		

		\begin{abstract}
			Photometric constraint is indispensable for self-supervised monocular depth estimation. It involves warping a source image onto a target view using estimated depth\&pose, and then minimizing the difference between the warped and target images. However, the endoscopic built-in light causes significant brightness fluctuations, and thus makes the photometric constraint unreliable. Previous efforts only mitigate this relying on extra models to calibrate image brightness. In this paper, we propose MonoPCC to address the brightness inconsistency radically by reshaping the photometric constraint into a cycle form. Instead of only warping the source image, MonoPCC constructs a closed loop consisting of two opposite forward-backward warping paths: from target to source and then back to target. Thus, the target image finally receives an image cycle-warped from itself, which naturally makes the constraint invariant to brightness changes. Moreover, MonoPCC transplants the source image's phase-frequency into the intermediate warped image to avoid structure lost, and also stabilizes the training via an exponential moving average (EMA) strategy to avoid frequent changes in the forward warping. The comprehensive and extensive experimental results on five datasets demonstrate that our proposed MonoPCC shows a great robustness to the brightness inconsistency, and exceeds other state-of-the-arts by reducing the absolute relative error by 7.27\%, 9.38\%, 9.90\% and 3.17\% on four endoscopic datasets, respectively; superior results on the outdoor dataset verify the competitiveness of MonoPCC for the natural scenario. 
			
		\end{abstract}
		
		\begin{keyword}
			Self-supervised learning \sep Monocular depth estimation \sep Photometric constraint \sep Brightness robustness
		\end{keyword}
		
	\end{frontmatter}
	
	
	\section{Introduction}
	
	Monocular endoscope is the key medical imaging tool for gastrointestinal diagnosis and surgery, but often provides a narrow field of view (FOV). 3D scene reconstruction helps enlarge the FOV, and also enables more advanced applications like surgical navigation by registration with preoperative computed tomography (CT). Depth estimation of monocular endoscopic images is prerequisite for reconstructing 3D structures, but extremely challenging due to absence of ground-truth (GT) depth labels. 
	
	\begin{figure}
	\centering
	\includegraphics[width=0.95\linewidth]{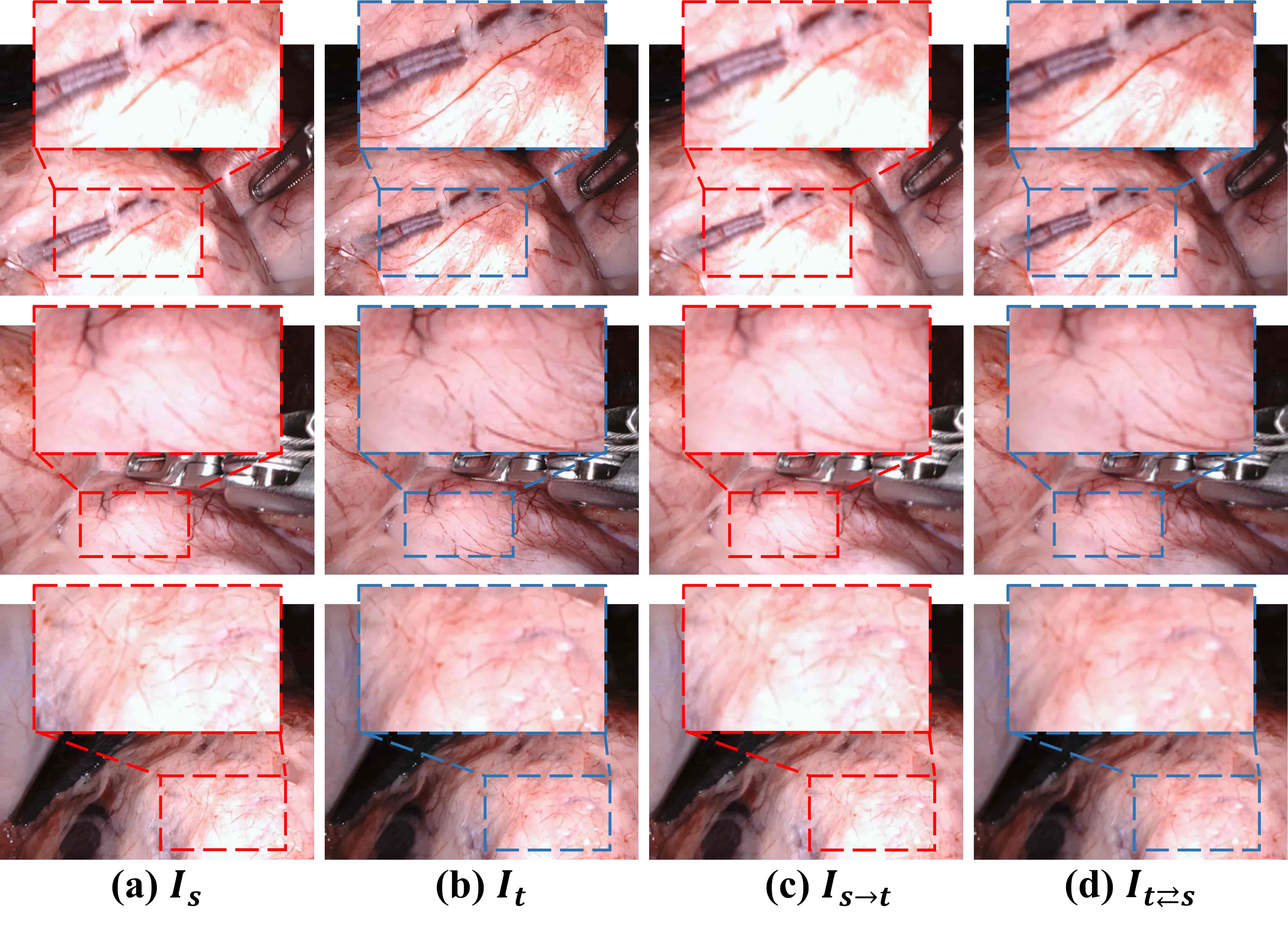}
	\caption{(a)-(b) are the source $I_{s}$ and target $I_{t}$ frames. (c) is the warped image from the source to target. (d) is the cycle-warped image along the target-source-target path for reliable photometric constraint. Box contour colors distinguish different brightness patterns.}
	\label{fig:introduction}
\end{figure}
	
	The typical solution of monocular depth estimation relies on self-supervised learning, where the core idea is photometric constraint between real and warped images. Specifically, two convolutional neural networks (CNNs) have to be built; one is called DepthNet and the other is PoseNet. The two CNNs estimate a depth map of each image and camera pose changes of every two adjacent images, based on which a source frame in an endoscopic video can be projected into 3D space and warped onto a target view of the other frame. DepthNet and PoseNet are jointly optimized to minimize a photometric loss, which is essentially the pixel-to-pixel difference between the warped and target images.

	However, the light source is fixed to the endoscope and moves along with the camera, resulting in significant brightness fluctuations between the source and target frames. Such problem can also be worsened by non-Lambertian reflection due to the close-up observation, as evidenced in Fig.~\ref{fig:introduction}(a)-(b).
	Consequently, between the target and warped source images, the brightness difference dominates as shown in Fig.~\ref{fig:introduction}(b)-(c), and thus misguides the photometric constraint in the self-supervised learning.
	
	Many efforts have been paid to enhance the reliability of photometric constraint under the fluctuated brightness. An intuitive solution is to calibrate the brightness in endoscopic video frames beforehand, using either a linear intensity transformation \citep{ozyoruk2021endoslam} or a trained appearance flow model \citep{shao2022self}. However, the former only addresses the global brightness inconsistency, and the latter increases the training difficulty due to introduced burdensome computation. Moreover, the reliability of appearance flow model is also not always guaranteed due to the weak self-supervision, which leads to a risk of wrongly modifying areas unrelated to brightness changes.
	
	In this paper, we aim to address the bottleneck of brightness inconsistency without relying on any auxiliary model. Our motivation stems from a recent method named TC-Depth \citep{ruhkamp2021attention}, which introduced a cycle warping originally for solving the occlusion issue. TC-Depth warps a target image to source and then warps back to itself to identify every occluded pixel, since they assume the occluded pixel can not come back to the original position precisely. We find that such cycle warping can naturally overcome the brightness inconsistency, and yield a more reliable warped image compared to only warping from source to target, as shown in Fig.~\ref{fig:introduction}(b)-(d). However, directly applying the cycle warping often fails in photometric constraint because (1) the twice bilinear interpolation of cycle warping blurs the image too much, and (2) the networks of depth\&pose estimation are learned actively, which makes the intermediate warping unstable and the convergence difficult.
	
	In view of the above analysis, we propose \textbf{Mono}cular depth estimation based on \textbf{P}hotometric-invariant \textbf{C}ycle \textbf{C}onstraint (MonoPCC), which adopts the idea of cycle warping but significantly reshapes it to enable the photometric constraint invariant to inconsistent brightness. Specifically, MonoPCC starts from the target image, and follows a closed loop path (target-source-target) to obtain a cycle-warped target image, which inherits consistent brightness from the original target image. To make such cycle warping effective in the photometric constraint, MonoPCC employs a learning-free structure transplant module (STM) based on Fast Fourier Transform (FFT) to minimize the negative impact of the blurring effect. STM restores the lost structure details in the intermediate warped image by `borrowing' the phase-frequency part from the source image. Moreover, instead of sharing the network weights in both target-source and source-target warping paths, MonoPCC bridges the two paths using an exponential moving average (EMA) strategy to stabilize the intermediate results in the first path.
	
	In summary, our main contributions are as follows:
	\begin{enumerate}
		\item We propose MonoPCC, which eliminates the inconsistent brightness inducing misguidance in the self-supervised learning by simply adopting a cycle-form warping to render the photometric constraint invariant to the brightness changes. 
		\item We introduce two enabling techniques, i.e., structure transplant module (STM) and an EMA-based stable training. STM restores the lost image details caused by interpolation, and EMA stabilizes the forward warping. These together guarantee an effective training of MonoPCC under the cycle-form warping. 
		\item We conduct comprehensive and extensive experiments on four public endoscopic datasets, i.e., SCARED \citep{allan2021stereo}, SimCol3D \citep{rau2023bimodal}, SERV-CT \citep{edwards2022serv}, and Hamlyn \citep{mountney2010three, stoyanov2010real, pratt2010dynamic} and a public natural dataset, i.e., KITTI \citep{geiger2012we}. The comparison results with eight state-of-the-art methods demonstrate the superiority of MonoPCC by decreasing the absolute relative error by 7.27\%, 9.38\%, 9.90\% and 3.17\% on four endoscopic datasets, respectively, and its strong ability to resist inconsistent brightness in the training. Moreover, the comparison results on KITTI further verify the competitiveness of MonoPCC even for the natural scenario, where the brightness changes are often not that significant.
	\end{enumerate}

	\section{Related Work}
	
	\subsection{Self-Supervised Depth Estimation}
	\label{SEC2.1}
	
	Self-supervised methods~\citep{zhou2017unsupervised,yin2018geonet,bian2019unsupervised,godard2019digging,watson2021temporal} of monocular depth estimation mostly, if not all, rely on the photometric constraint to train two networks, named DepthNet and PoseNet. 
	For example, as a pioneering work, SfMLearner \citep{zhou2017unsupervised} estimated depth maps and camera poses to synthesize a fake target image by warping another image from the source view, and calculated a reconstruction loss between the warped and target images as the photometric constraint.  
	
	For endoscopic cases, M3Depth \citep{huang2022self} enforced not only the photometric constraint but also a 3D geometric structural consistency by Mask-ICP. For improving pose estimation in laparoscopic procedures, \citet{li2022geometric} designed a Siamese pose estimation scheme and also constructed dual-task consistency by additionally predicting scene coordinates. These methods share a common flaw: brightness fluctuations caused by the moving light source in the endoscope, significantly reduce the effectiveness of photometric constraint in self-supervised learning, degrading the performance consequently.
	
	\begin{figure*}[t]
	\centering
	\includegraphics[width=0.95\linewidth]{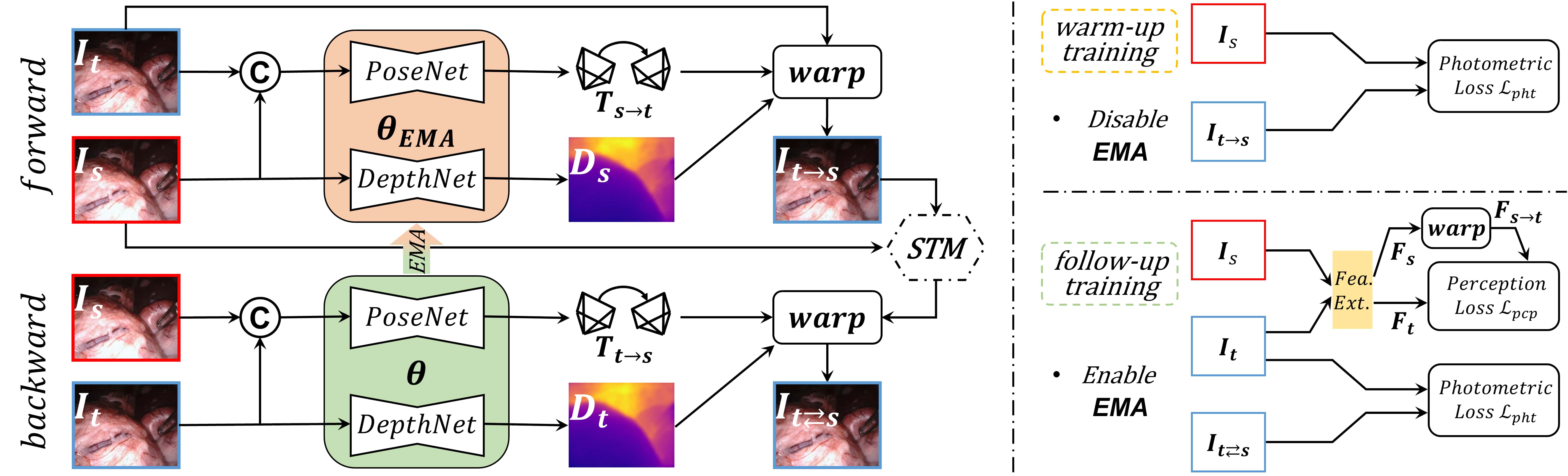}
	\caption{The training pipeline of MonoPCC, which consists of forward and backward cascaded warping paths bridged by two enabling techniques, i.e., structure transplant module (STM) and exponential moving average (EMA). The training has two phases, i.e., warm-up to initialize the network weights for reasonable forward warping, and follow-up to resist the brightness changes. Different box contour colors code different brightness patterns. $\copyright$ means concatenation.}
	\label{fig:pipeline}
\end{figure*}
	
	\subsection{Brightness Inconsistency}
	\label{SEC2.2}
	Recently, several efforts have been made to mitigate brightness inconsistency in endoscopic cases, the common solution is to calibrate the brightness inconsistency using either preprocessing or extra learned models. For example, Endo-SfMLearner \citep{ozyoruk2021endoslam} calculated the mean values and standard deviation of images, and then linearly aligned the brightness of the warped image with that of the target one. However, the real brightness fluctuations are actually local and non-linear, and are hardly calibrated by a simple linear transformation. Recently, AF-SfMLearner \citep{shao2022self} tried to relax the requirement of brightness consistency by learning an appearance flow model, which estimates pixel-wise brightness offset between adjacent frames. Nevertheless, effectiveness relies on the quality of the learned appearance flow model, and the risk of wrongly modifying the image content occurs accompanying the model error. Also, introducing an extra model dramatically increases the training time and difficulty in self-supervised learning.

	\subsection{Cycle Warping in Medical Image Analysis}
	\label{SEC2.3}	
	Cycle warping is a crucial technique in the context of image processing, and its applications have been explored in various domains. The core concept behind cycle constraint is to apply the same operation twice in opposite directions, ensuring cycle consistency between input and output results. Early work on this concept was applied in dense correspondence tasks \citep{zhou2016learning}, where a 3D computer aided design (CAD) model was used to generate two synthetic views of paired images, and cycle consistency was enforced between these two synthetic views. Since then, cycle constraint has been extended to other tasks, such as video interpolation \citep{reda2019unsupervised}, object tracking \citep{yuan2020self}, and label propagation \citep{ganeshan2021warp}. More recently, cycle constraint has been employed in medical imaging scenarios with impressive results \citep{kim2021cyclemorph,jang2023unsupervised,lu2023two,gong2024self}. For example, to preserve original topology during deformable registration, CycleMorph \citep{kim2021cyclemorph} trained two registration networks to generate forward and reverse directional deformation vector fields, yielding the cycle consistency between the reversed image and the original image. \citet{gong2024self} introduced a cycle constraint-based approach for modeling surgical soft-tissue deformation. By measuring the similarity between warped pixels throughout the entire forward-backward cycle, this method demonstrated robust deformation modeling across various tissue patterns, making it particularly useful for surgical planning and simulation. However, directly applying cycle warping to our specific task would lead to significant side effects, such as loss of detail and increased training difficulty, due to the twice-warping operation.

	\subsection{Fast Fourier Transform in Medical Image Analysis}
	\label{SEC2.4}
	The Fast Fourier Transform (FFT) is a powerful technique that converts signals from the spatial domain to the frequency domain and has been widely applied in medical imaging tasks. For instance, FFT is commonly used in CT reconstruction \citep{shepp1974fourier} and MRI denoising \citep{wink2004denoising}. Recent advances in deep learning have further extended the application of FFT to other areas, particularly for addressing domain shifts, such as domain generalization \citep{liu2021feddg,zhao2024morestyle}, domain adaptation \citep{yang2022source,wang2023curriculum}, and style transfer \citep{lv2023robust}. In the context of domain generalization, FedDG \citep{liu2021feddg} was developed to improve the generalizability of federated learning models across unseen domains. By extracting low-level distribution information in the frequency space of client samples and performing continuous interpolation between local and diverse domain samples, FedDG enables each local client to acquire information from multiple domains, enhancing the model’s robustness. To alleviate the spatial misalignment in atlas-based segmentation, \citet{lv2023robust} proposed an image-aligned style transformation to generate image-mask pairs with diverse patterns. Specifically, Fourier amplitude components between the target image and warped atlas were mixed, while the phase component of the warped atlas was retained. Inspired by \citet{lv2023robust}, we align source-target image pairs before Structure Transplant Module (STM), thereby mitigating the risk of potential artifacts. Conversely, we ensure the preservation of the brightness pattern by retaining the amplitude components while swapping the phase components to reconstruct blurred details caused by warping.

	\section{Method}
	
	Fig.~\ref{fig:pipeline} illustrates the pipeline of MonoPCC, consisting of both forward and backward warping paths in the training phase. 
	We first explain how to warp images for self-supervised learning in Sec.~\ref{SEC3.1}, and then detail the photometric-invariant principle of MonoPCC in Sec.~\ref{SEC3.2}, as well as its two key enabling techniques in Sec.~\ref{SEC3.3} and Sec.~\ref{SEC3.4}, i.e., structure transplant module (STM) for avoiding detail lost and EMA between two paths for stabilizing the training.

	\subsection{Warping across Views for Self-Supervision}
	\label{SEC3.1}
	
	To warp a source image $I_{s}$ to a target view $I_{t}$, DepthNet and PoseNet first estimate the target depth map $D_{t}$, and the camera pose changing from target to source $T_{t \rightarrow s}$, respectively.
	
	The DepthNet is typically an encoder-decoder, which inputs a single endoscopic image and outputs an aligned depth map. In this work, we use MonoViT \citep{zhao2022monovit} as our backbone DepthNet. The PoseNet contains a lightweight ResNet18-based \citep{he2016deep} encoder with four convolutional layers, which inputs a concatenated two images and predicts the pose change from the image in the tail channels to that in the front channels. 
	
	Using $D_t$ and $T_{t \rightarrow s}$, the warping can be performed from the source to target view, and each pixel in the warped image $I_{s \rightarrow t}$ can find its matching pixel in the source image using the following equation:
		\begin{equation}
			p_s = K T_{t \rightarrow s} D_t(p_{s \rightarrow t}) K^{-1} p_{s \rightarrow t},
			\label{equ:warp}
		\end{equation}
	where $p_s$ and $p_{s \rightarrow t}$ are the pixel's homogeneous coordinates in $I_{s}$ and $I_{s \rightarrow t}$, respectively, $D(p)$ means the depth value of $D$ at the position $p$, and $K$ denotes the given camera intrinsic matrix. With the pixel matching relationship, $I_{s \rightarrow t}$ can be obtained by filling the color at each pixel position using differentiable bilinear sampling \citep{jaderberg2015spatial}:
	\begin{equation}
		I_{s \rightarrow t}(p_{s \rightarrow t})=\mathtt{BilinearSampler}(I_s(p_s)),
		\label{equ:bisample}
	\end{equation}
	where $I(p)$ means the pixel intensity of $I$ at the position $p$. Since $p_{s \rightarrow t}$ is discrete and $p_{s}$ is continuous, $\mathtt{BilinearSampler}$ is utilized to calculate the intensity of each pixel in $I_{s \rightarrow t}$ using the neighboring pixels around $p_{s}$ and allow error backpropagation. 
	
	After warping, DepthNet and PoseNet can be constrained via the photometric loss, which is calculated as follows: 
		\begin{equation}
			\begin{split}
				\mathcal{L}_{pht}(I_t, I_{s \rightarrow t}) &= 
				\alpha \cdot \frac{1-\operatorname{SSIM}(I_t, I_{s \rightarrow t})}{2} \\
				&+ (1-\alpha) \cdot|I_t-I_{s \rightarrow t}|,
				\label{equ:l1ssim}
			\end{split}
		\end{equation}
	where Structure Similarity Index Measure (SSIM) \citep{wang2004image} constrains the image quality and L1 loss constrains the content. $\alpha$ is set to 0.85 according to \citet{godard2017unsupervised}.
	
	One of the assumptions for the photometric constraint in Eq.~(\ref{equ:l1ssim}) is that the scene contains no specular reflections, and is temporally consistent in terms of brightness \citep{shao2022self}. However, such assumption hardly holds true in the endoscopic cases as visualized in Fig.~\ref{fig:introduction}.

	\subsection{Photometric-invariant Cycle Warping}
	\label{SEC3.2}
	
	To overcome the brightness inconsistency, we construct a cycle loop path involving forward and backword warping, as shown in Fig.~\ref{fig:pipeline}. To better explain the principle, we use the red and blue box contours to indicate the different brightness patterns carried by the image.
	
	To get a warped image inheriting the target's brightness, we start from the target itself, and first warp it to get $I_{t \rightarrow s}$, and then warp back to get $I_{t \rightleftarrows s}$. The pixel matching relationships across the three images are formulated as follows:
	\begin{equation}
		\left\downarrow
		\begin{aligned}
			&p_{t} = K T_{s \rightarrow t} D_s(p_{t \rightarrow s}) K^{-1} p_{t \rightarrow s}, \\
			&p_{t \rightarrow s} = K T_{t \rightarrow s} D_t(p_{t \rightleftarrows s}) K^{-1} p_{t \rightleftarrows s},
		\end{aligned}
		\right.
		\label{equ:cycle_warp}
	\end{equation}
	where $p_t$, $p_{t \rightarrow s}$, and $p_{t \rightleftarrows s}$ are the pixel's homogeneous coordinates in $I_{t}$, $I_{t \rightarrow s}$ and $I_{t \rightleftarrows s}$, respectively. $D_{s}$ and $D_{t}$ are DepthNet-predicted depth maps of the source and target images, respectively.
	
	Therefore, according to Eq.~(\ref{equ:bisample}), $I_{t \rightarrow s}$ and $I_{t \rightleftarrows s}$ can be obtained sequentially:
	\begin{equation}
		\left\downarrow
		\begin{aligned}		
			&I_{t \rightarrow s} (p_{t \rightarrow s}) = \mathtt{BilinearSampler}(I_{t}(p_{t})),\\
			&I_{t \rightleftarrows s} (p_{t \rightleftarrows s}) = \mathtt{BilinearSampler}(I_{t \rightarrow s}(p_{t \rightarrow s})).
		\end{aligned}
		\right.
		\label{equ:cycle_sample}
	\end{equation}
	
	Thus, we rewrite the previous photometric constraint Eq.~(\ref{equ:l1ssim}) to a cycle form, i.e., $\mathcal{L}_{pht}(I_t, I_{t \rightleftarrows s})$, where $I_{t}$ and $I_{t \rightleftarrows s}$ have the same brightness pattern as shown in Fig.~\ref{fig:pipeline}. However, direct usage of such cycle warping brings two issues in the optimization of photometric constraint: (1) noticeable image detail lost due to twice image interpolation using $\mathtt{BilinearSampler}$, which negatively affects the appearance-based photometric constraint, and (2) the networks learn too actively to give stable intermediate warping, making the training of two networks difficult to converge. Thus, we introduce two enabling techniques, i.e., structure transplant module to retain structure details and EMA strategy to stabilize forward warping.

	\begin{figure}[t]
		\centering
		\includegraphics[width=0.9\linewidth]{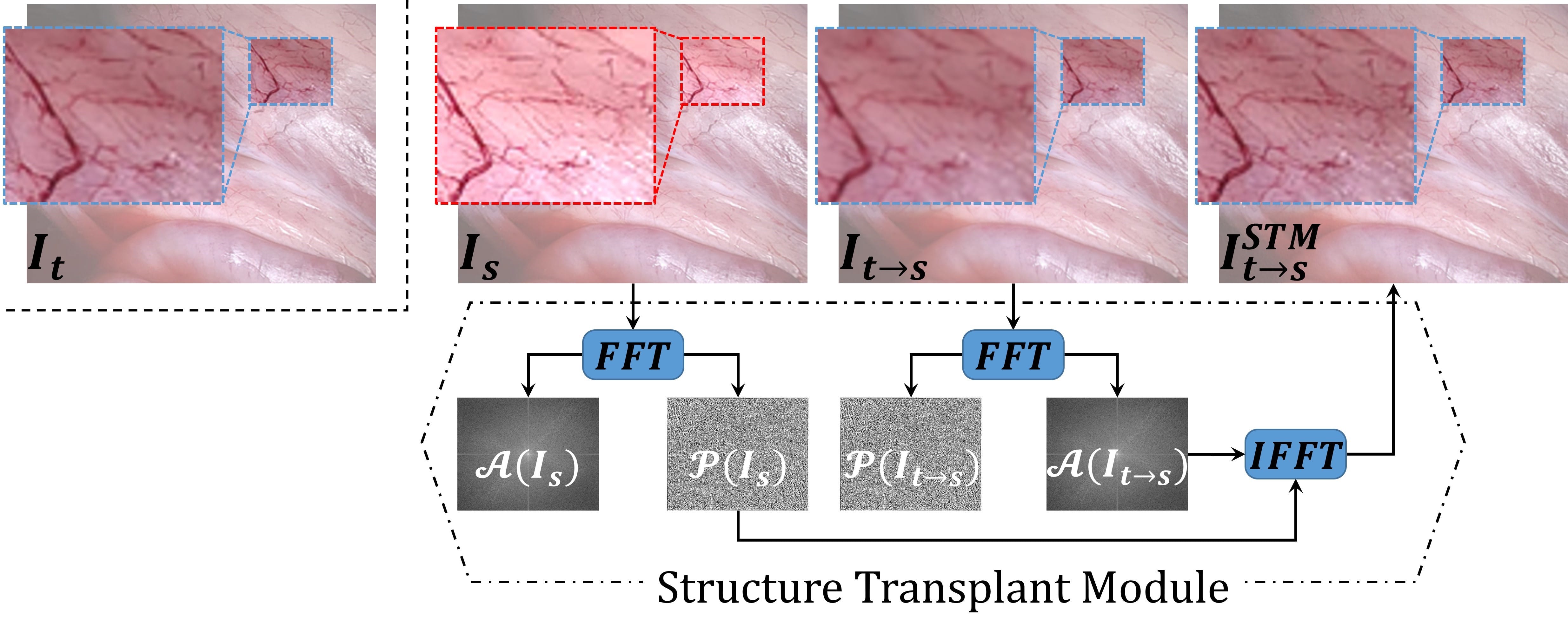}
		\caption{Details of STM, which utilizes the phase-frequency of the source image $I_{s}$ to replace that of the warped image $I_{t \rightarrow s}$ to avoid image detail lost.
		}
		\label{fig:comparison_STM}
	\end{figure}
	
	\subsection{Structure Transplant to Retain Details}
	\label{SEC3.3}

	The learning-free structure transplant module (STM) works based on two observations: (1) the warped frame $I_{t \rightarrow s}$ has homologous image style as $I_t$, and (2) the original frame $I_s$ has real structure details, and also is roughly aligned with $I_{t \rightarrow s}$. Inspired by the image-aligned style transformation proposed in \citet{lv2023robust}, we transplant fine structure of $I_s$ onto the appearance of $I_{t \rightarrow s}$ via a learning-free style transformation approach illustrated in Fig.~\ref{fig:comparison_STM}. 
	
	Specifically, STM utilizes Fast Fourier Transform (FFT) \citep{nussbaumer1982fast} to decompose an RGB image $x$ into two components, i.e., amplitude $\mathcal{A}(x)$ and phase $\mathcal{P}(x)$. $\mathcal{A}(x)$ mainly controls the image style like brightness and $\mathcal{P}(x)$ contains the information of structure details \citep{lv2023robust}. Thus, after the forward warping, STM recombines $\mathcal{A}(I_{t \rightarrow s})$ and $\mathcal{P}(I_s)$, and produces the structure-restored warped image $I_{t \rightarrow s}^{STM}$ via Inverse Fast Fourier Transform (IFFT):  
		\begin{equation}
			I_{t \rightarrow s}^{STM} = IFFT(\mathcal{A}(I_{t \rightarrow s}) * e^{-j * \mathcal{P}(I_s)} ).
			\label{equ:stm}
		\end{equation}
	
	Therefore, in the backward warping, we utilize $I_{t \rightarrow s}^{STM}$ to substitute $I_{t \rightarrow s}$ in Eq.~(\ref{equ:cycle_sample}), which can be rewritten as follows:
		\begin{equation}
			I_{t \rightleftarrows s} (p_{t \rightleftarrows s}) = \mathtt{BilinearSampler}(I_{t \rightarrow s}^{STM} (p_{t \rightarrow s})).
			\label{equ:stm_sample}
		\end{equation}
	
	As shown in Fig.~\ref{fig:comparison_STM}, $I_{t \rightarrow s}^{STM}$ displays better structure details than $I_{t \rightarrow s}$, and also maintains similar illumination as $I_{t \rightarrow s}$.

	\subsection{EMA to Stabilize Forward Warping}
	\label{SEC3.4}
	
	As shown in Fig.~\ref{fig:pipeline}, the two paths are connected in a cascaded fashion, where the result of the second warping relies on the output of the previous warping. Therefore, a well-initialized and steady intermediate warped image $I_{t \rightarrow s}$ is very important for a stable convergence of training. In view of this, we introduce an exponential moving average (EMA) to bridge the two paths and thus divide the whole training into warm-up and follow-up phases.
	
	During the warm-up training, EMA is disabled and we only train DepthNet and PoseNet in the forward warping by optimizing the photometric constraint between $I_{s}$ and $I_{t \rightarrow s}$, that is, $\theta = \arg\min_{\theta} \mathcal{L}_{pht}(I_{s}, I_{t \rightarrow s})$, where $\theta$ represents the total learnable parameters of both DepthNet and PoseNet. Although the brightness inconsistency occurs in the warm-up training, a reasonably acceptable initialization of DepthNet and PoseNet can be reached. 
	
	\begin{figure}[t]
		\centering
		\includegraphics[width=0.9\linewidth]{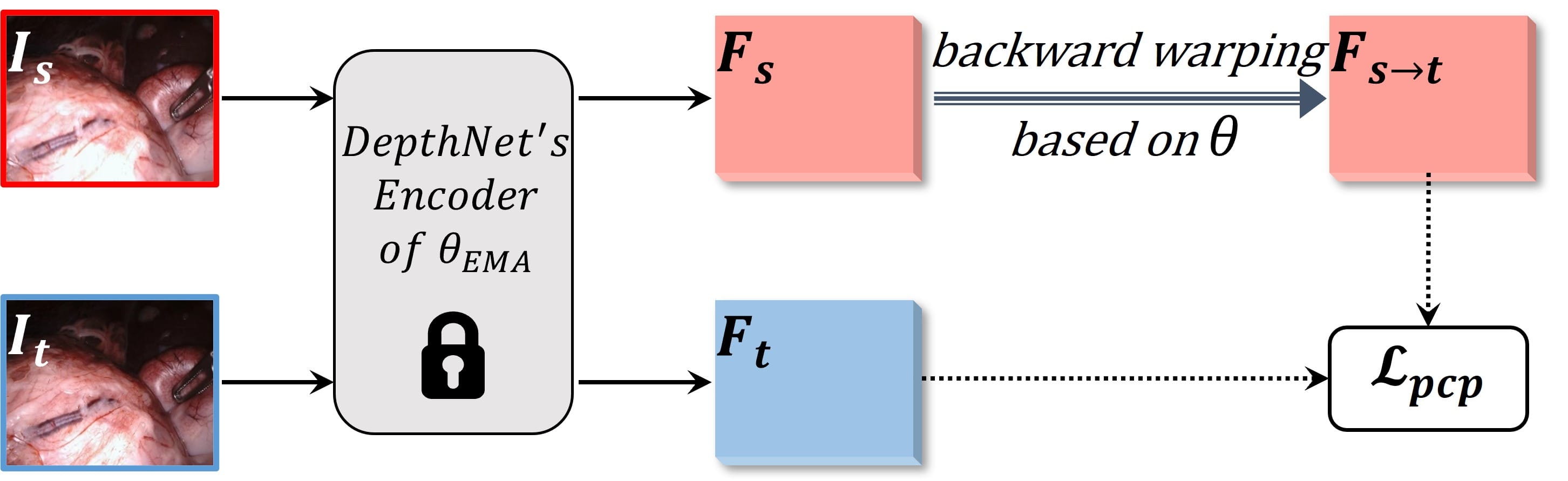}
		\caption{The auxiliary perception constraint by backward warping the encoding feature maps instead of raw images.
		}
		\label{fig:perceptual}
	\end{figure}
	
	The follow-up training with EMA enabled is the key step to resist brightness inconsistency based on the cycle warping. We duplicate the network parameters learned in the warm-up training as an EMA copy $\theta_{EMA} \leftarrow \theta$, and use the models with $\theta_{EMA}$ to estimate depth and pose in the forward warping and those with $\theta$ in the backward warping. $\theta$ is updated actively by minimizing the cycle-form photometric loss $L_{pht}(I_{t},I_{t \rightleftarrows s})$, and $\theta_{EMA}$ is updated slowly by moving averaging $\theta$. That is, the copy of DepthNet and PoseNet for the forward warping is \textbf{not} directly learned by the optimization of photometric constraint in the follow-up training, thus to avoid predicting frequently changed $I_{t \rightarrow s}$. 
	
	Besides, we further enhance the learning robustness via a ready-made perception loss. As shown in Fig.~\ref{fig:perceptual}, we backward warp the encoding map of $I_{s}$ generated from DepthNet's encoder of its EMA version, yielding a warped feature map $F_{s \rightarrow t}$. If we also extract the encoding map of $I_{t}$ using the same encoder, $F_{s \rightarrow t}$ and $F_{t}$ should be aligned and may contain high-level semantic information covering the low-level brightness changes. Thus, a perception loss is defined:
		\begin{equation}
			\mathcal{L}_{pcp}(F_t, F_{s \rightarrow t}) = |F_t - F_{s \rightarrow t}|.
			\label{equ:Lpcp}
		\end{equation}
	
	In summary, the follow-up training of MonoPCC is formulated as follows:
	\begin{equation}
		\left\{
		\begin{aligned}		
			&\theta = \arg\min_{\theta} (\mathcal{L}_{pht}(I_{t}, I_{t \rightleftarrows s}) + \mathcal{L}_{pcp}(F_t, F_{s \rightarrow t})),\\
			&\theta_{EMA} = \alpha \theta_{EMA} + (1 - \alpha) \theta,
		\end{aligned}
		\right.
	\end{equation}
	where $\alpha$ is a hyperparameter to balance the stability and learnability of the forward warping, and set to $0.75$ in MonoPCC by default. 
	
		\begin{table}
		\small
		\caption{Evaluation metrics of monocular depth estimation, where $N$ refers to the number of valid pixels in depth maps, $d_i$ and $d_i^*$ denote the estimated and GT depth of $i$-th pixel, respectively. The Iverson bracket $[\cdot]$ yields 1 if the statement is true, otherwise 0.}
		\begin{center}
			\begin{tabular}{ cc }
				\toprule
				Metric         &  Formula \\
				\midrule
				Abs Rel    &  $\frac{1}{N} {\sum\limits_{i = 0}^{N - 1} {|d_i - d_i^*|} / {d_i^*} }$ \\ 
				Sq Rel    &  $\frac{1}{N} {\sum\limits_{i = 0}^{N - 1} {|d_i - d_i^*|^2} / {d_i^*} }$ \\ 
				RMSE       &  $\sqrt{\frac{1}{N} {\sum\limits_{i = 0}^{N - 1} {|d_i - d_i^*|}^2 }}$ \\ 
				RMSE log   &  $\sqrt{\frac{1}{N} {\sum\limits_{i = 0}^{N - 1} {|log\,d_i - log\,d_i^*|}^2 }}$ \\
				$\delta$       &  $\frac{1}{N} \sum\limits_{i = 0}^{N - 1}  [max(\frac{d_i}{d_i^*}, \frac{d_i^*}{d_i})\, < \, 1.25] $ \\
				\bottomrule
			\end{tabular}
			
			\label{tab:metrics}
		\end{center}
	\end{table}	
	
	\subsection{Implementation Details}
	\label{SEC3.5}
	
	We train MonoPCC using a single NVIDIA RTX A6000 GPU. We utilize MonoViT and ResNet18 pre-trained on ImageNet\citep{deng2009imagenet} for DepthNet and PoseNet initialization, respectively. 
	We set the batch size to 12, and use AdamW \citep{loshchilov2017decoupled} as the optimizer. The learning rate decay is exponential and set to 0.9. The number of epochs and learning rate differ in the warm-up and follow-up training phases. In warm-up training, the number of epochs is set to 20 epochs, and the initial learning rate is set to $1 \times 10^{-4}$ for DepthNet's encoder and $5 \times 10^{-5}$ for the rest weights. In follow-up training, the number of epochs is set to 10 epochs, and the initial learning rate is set to $5 \times 10^{-5}$ for all network weights. The weights of estimation networks in the EMA-copy path are updated every 200 steps.

	\section{Experimental Settings}
	
	\subsection{Comparison Methods}
	\label{SEC4.1}
	
	We compare MonoPCC with 8 state-of-the-art (SOTA) methods of monocular depth estimation, i.e., Monodepth2 \citep{godard2019digging}, FeatDepth \citep{shu2020feature}, HR-Depth \citep{lyu2021hr}, DIFFNet \citep{zhou2021self}, Endo-SfMLearner \citep{ozyoruk2021endoslam}, AF-SfMLearner \citep{shao2022self}, MonoViT \citep{zhao2022monovit}, and Lite-Mono \citep{zhang2023lite}. 
	We evaluate their performance and make comparisons using their released codes.

	\begin{table*}[]
		\small
		\centering
		\caption{Quantitative comparison results on SCARED and SimCol3D. The best results are marked in bold and the second-best underlined. The paired \textit{p}-values between MonoPCC and others are all less than $0.05$. }
		\begin{tabular}{lccccc|ccccc}
			\toprule
			\multirow{2}{*}{Methods} & \multicolumn{5}{c|}{SCARED} & \multicolumn{5}{c}{SimCol3D} \\
			\cmidrule(r){2-6} \cmidrule(r){7-11}
			& Abs Rel $\downarrow$ & Sq Rel $\downarrow$ & RMSE $\downarrow$  & RMSE log $\downarrow$ & $\delta$ $\uparrow$ & Abs Rel $\downarrow$ & Sq Rel $\downarrow$ & RMSE $\downarrow$  & RMSE log $\downarrow$ & $\delta$ $\uparrow$    \\
			\midrule
			Monodepth2      & 0.060          & 0.432          & 4.885          & 0.082          & 0.972          & 0.076          & 0.061          & 0.402          & 0.106          & 0.950           \\
			FeatDepth       & \underline{0.055}          & 0.392          & 4.702          & 0.077          & 0.976          & 0.077          & 0.069          & 0.374          & 0.098          & 0.957           \\
			HR-Depth        & 0.058          & 0.439          & 4.886          & 0.081          & 0.969          & 0.072          & 0.044          & 0.378          & 0.100          & 0.961           \\
			DIFFNet         & 0.057          & 0.423          & 4.812          & 0.079          & 0.975          & 0.074          & 0.053          & 0.401          & 0.105          & 0.957           \\
			Endo-SfMLearner & 0.057          & 0.414          & 4.756          & 0.078          & 0.976          & 0.072          & 0.042          & 0.407          & 0.103          & 0.950           \\
			AF-SfMLearner   & \underline{0.055}          & \underline{0.384}  & \underline{4.585}  & \underline{0.075}  & \underline{0.979}  & 0.071          & 0.045          & \underline{0.372}  & 0.099          & 0.961           \\
			MonoViT         & 0.057          & 0.416          & 4.919          & 0.079          & 0.977          & \underline{0.064}  & \underline{0.034}  & 0.377          & \underline{0.094}  & \underline{0.968}   \\
			Lite-Mono   & 0.056          & 0.398          & 4.614          & 0.077          & 0.974          & 0.076          & 0.050          & 0.424          & 0.110          & 0.950           \\
			MonoPCC(Ours)   & \textbf{0.051} & \textbf{0.349} & \textbf{4.488} & \textbf{0.072} & \textbf{0.983} & \textbf{0.058} & \textbf{0.028} & \textbf{0.347} & \textbf{0.090} & \textbf{0.975} \\
			\bottomrule
		\end{tabular}
		
		\label{tab:scared_simcol}
	\end{table*}

	\begin{figure*}[t]
		\centering
		\includegraphics[width=1.0\linewidth]{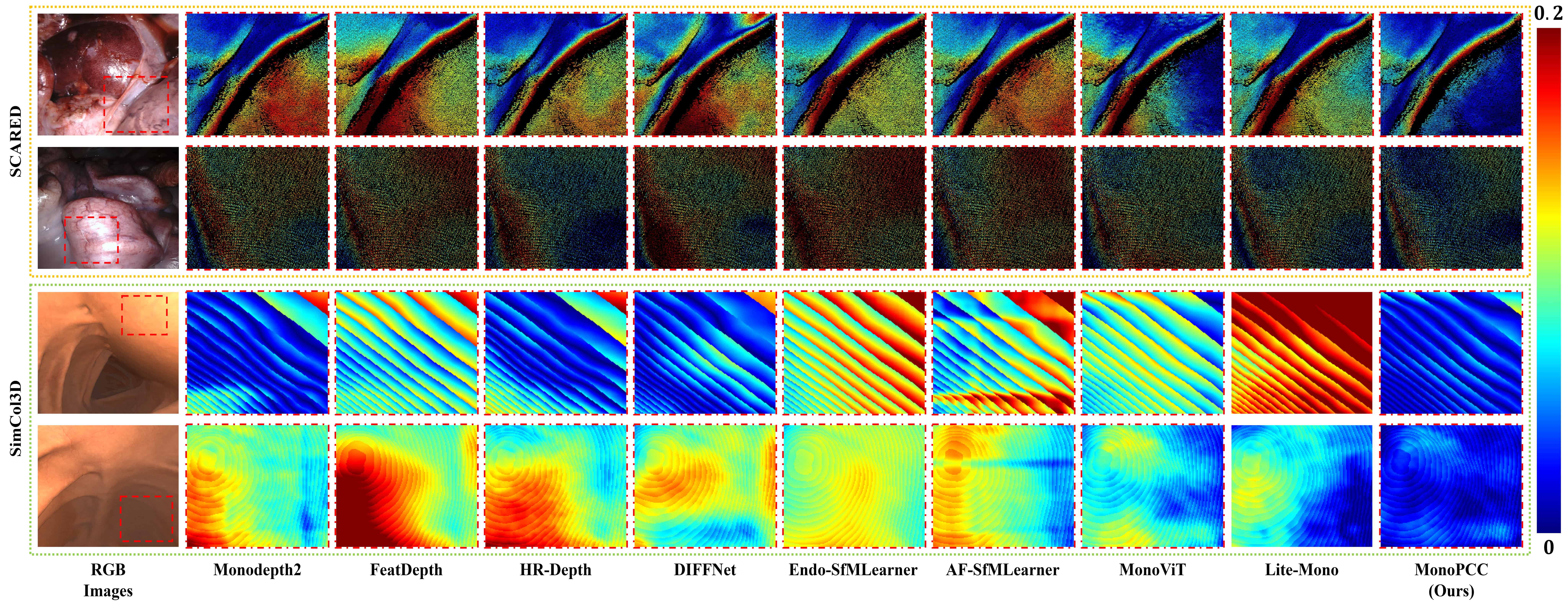}
		\caption{The Abs Rel error maps of comparison methods on SCARED and SimCol3D, with close-up details highlighted. The regions of interest (ROIs) are outlined with red dashed lines, and the Opencv Jet Colormap is used for visualization. }
		\label{fig:scared_simcol}
	\end{figure*}

	\subsection{Datasets}
	\label{SEC4.2}
	
	The experiment in this work involves 4 public endoscopic datasets, i.e., SCARED \citep{allan2021stereo}, SimCol3D \citep{rau2023bimodal}, SERV-CT \citep{edwards2022serv}, and Hamlyn \citep{mountney2010three, stoyanov2010real, pratt2010dynamic}, and a public outdoor-scene dataset, i.e., KITTI \citep{geiger2012we}. 
	
	\textbf{SCARED} is from a MICCAI challenge and consists of 35 videos with the size of $1280 \times 1024$ collected from porcine cadavers using a da Vinci Xi surgical robot. 
	Structured light is used to obtain the ground-truth (GT) depth map for the first frame, which is then mapped onto the subsequent frames using camera trajectory recorded by the robotic arm.
	
	Since the original data is binocular, we only use the left view to simulate our focusing monocular situation. We split the dataset into 21,066 frames (from 28 videos) for training and 551 (from the rest 7 videos) for test, which are consistent with the previous SOTA \citep{shao2022self}. No videos are cross-used in both training and test. Also, two trajectories from the 7 videos are used to evaluate the performance of pose estimation, including 410 and 833 frames, respectively.

	\textbf{SimCol3D} is from MICCAI 2022 EndoVis challenge. It is a synthetic dataset which contains over 36,000 colonoscopic images and depth annotations with size of $475 \times 475$. Meanwhile, virtual light sources are attached to the camera. For the usage of SimCol3D dataset, we follow their official website
	and spilt the dataset into 28,776 and 9,009 frames for training and test, respectively. 
	
	\textbf{SERV-CT} is collected from two \textit{ex vivo} porcine cadavers, and each has 8 binocular keyframes. We treat the two views independently and thus have 32 images in total with the size of $720 \times 576$. The GT depth map for each image is calculated by manually aligning the endoscopic image to the 3D anatomical model derived from the corresponding CT scan.
	
	\textbf{Hamlyn} is a large public endoscopic dataset with multiple stereo videos for clinical application, including roughly 8,000 frames with different resolutions. \citet{recasens2021endo} provided a rectified version and created GT depth maps for left-view frames using software \textit{Libelas}.  
	
	\textbf{KITTI} is an outdoor dataset captured in autonomous driving scenarios, e.g., road and campus, while laser sensors are equipped to scan precise depth maps. We follow the split of \citet{eigen2014depth} and \citet{zhou2017unsupervised}, and use 39,810 frames for training, 4,424 for validation, and 697 for test.

	\subsection{Evaluation Metrics}
	\label{SEC4.3}
	
	To be consistent with previous works, we employ 5 metrics in the evaluation, which are listed in Table \ref{tab:metrics}. 
	
	Note that, as the common evaluation routine of monocular depth estimation, the predicted depth map should be scaled beforehand since it is scale unknown. The scaling factor is the GT median depth divided by the predicted median depth \citep{zhou2017unsupervised}. 
	%
	
	Also, for a fair comparison, we follow the previous works \citep{shao2022self,rau2023bimodal,recasens2021endo,godard2019digging}, and cap the depth maps within the maximum value for each dataset. Specifically, the maximum depth value is set to $150\ mm$, $200\ mm$, $180\ mm$, $300\ mm$, and $80\ m$ for SCARED, SimCol3D, SERV-CT, Hamlyn and KITTI, respectively.

	To evaluate the performance of camera pose estimation, we follow a protocol similar to that used for depth estimation. Specifically, we align the predicted camera poses to the scale of the ground truth trajectory, and we adopt the 5-frame evaluation rule \citep{zhou2017unsupervised}, using the absolute trajectory error (ATE) as the evaluation metric.

	Furthermore, for each comparison in the following experimental results, we perform paired samples T-test \citep{duncan1975t} and report the \textit{p}-value to indicate whether the difference is significant.

	\section{Results and Discussions}
	\label{sec:results}
	
	\subsection{Comparison with State-of-the-arts}
	\label{SEC5.1}
	
	\begin{table*}[]
		\small
		\centering
		\caption{Quantitative comparison results on SERV-CT and Hamlyn. The best results are marked in bold and the second-best underlined. The paired \textit{p}-values between MonoPCC and others are all less than $0.05$, except for $\delta$ on SERV-CT compared to MonoViT.}
		{\begin{tabular}{lccccc|ccccc}
				\toprule
				\multirow{2}{*}{Methods} & \multicolumn{5}{c|}{SERV-CT} & \multicolumn{5}{c}{Hamlyn} \\
				\cmidrule(r){2-6} \cmidrule(r){7-11}
				& Abs Rel $\downarrow$ & Sq Rel $\downarrow$ & RMSE $\downarrow$  & RMSE log $\downarrow$ & $\delta$ $\uparrow$ & Abs Rel $\downarrow$ & Sq Rel $\downarrow$ & RMSE $\downarrow$  & RMSE log $\downarrow$ & $\delta$ $\uparrow$    \\
				\midrule
				Monodepth2      & 0.127 & 2.152 & 13.023 & 0.166 & 0.825           & 0.087          & 1.254          & 9.555          & 0.115          & 0.939           \\
				FeatDepth       & 0.117 & 1.862 & 12.040 & 0.154 & 0.841           & 0.064          & 0.939          & 8.016          & \underline{0.090}  & 0.964   \\
				HR-Depth        & 0.122 & 2.085 & 12.587 & 0.156 & 0.850           & 0.076          & 1.139          & 8.883          & 0.102          & 0.957           \\
				DIFFNet         & 0.116 & 1.858 & 12.177 & 0.146 & 0.864           & 0.070          & 1.093          & 8.552          & 0.097          & 0.958           \\
				Endo-SfMLearner & 0.122 & 2.123 & 12.551 & 0.168 & 0.842           & \underline{0.063}  & \underline{0.886}  & \underline{7.901}  & 0.091          & 0.968           \\
				AF-SfMLearner   & \underline{0.101} & \underline{1.546} & \underline{10.900} & \underline{0.131} & 0.888            & 0.078          & 1.018          & 8.712          & 0.104          & 0.968           \\
				MonoViT         & 0.103 & 1.566 & 11.482 & 0.136 & \underline{0.895}         & 0.073          & 0.945          & 8.470          & 0.099          & \underline{0.975}           \\
				Lite-Mono   & 0.124 & 2.314 & 13.156 & 0.175 & 0.820           & 0.074          & 1.106          & 8.743          & 0.104          & 0.950           \\
				MonoPCC(Ours)   & \textbf{0.091} & \textbf{1.252} & \textbf{10.059} & \textbf{0.116} & \textbf{0.915}             & \textbf{0.061} & \textbf{0.819} & \textbf{7.604} & \textbf{0.085} & \textbf{0.977} \\
				\bottomrule
		\end{tabular}}
		
		\label{tab:generalization}
	\end{table*}

	\begin{figure*}[t]
		\centering
		\includegraphics[width=1.0\linewidth]{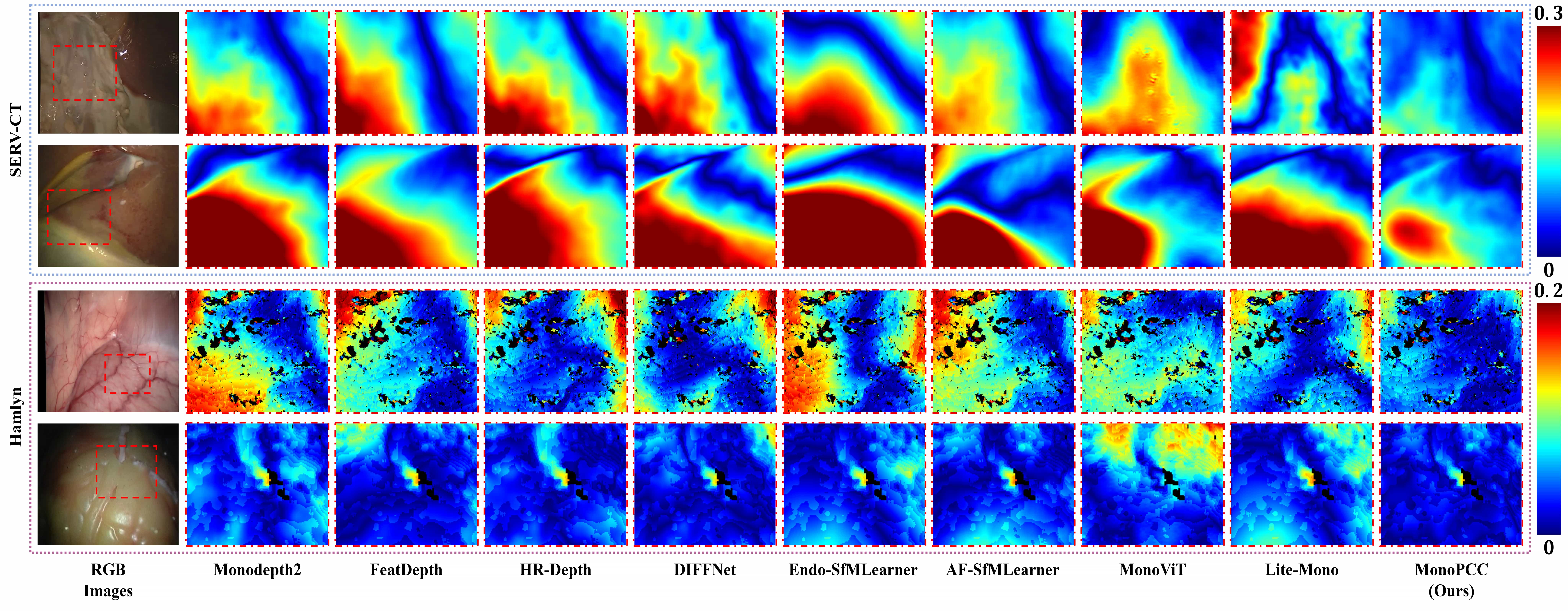}
		\caption{The Abs Rel error maps of comparison methods on SERV-CT and Hamlyn, with close-up details highlighted. The regions of interest (ROIs) are outlined with red dashed lines, and the Opencv Jet Colormap is used for visualization. }
		\label{fig:ct_hamlyn}
	\end{figure*}

	\subsubsection{Evaluation on SCARED and SimCol3D}
	\label{scared_shao}
	Table~\ref{tab:scared_simcol} shows the comparison results on SCARED (left part) and SimCol3D (right part). As can be seen, our method achieves the best performance in terms of all five metrics on SCARED, and consistently outperforms the other SOTAs by different margins on SimCol3D.
	
	Likewise, Endo-SfMLearner and AF-SfMLearner also belong to the approach of addressing the brightness inconsistency in self-supervised monocular depth estimation. Endo-SfMLearner adopts the simple idea of brightness linear transformation, while AF-SfMLearner resorts to a more complex appearance flow model. 
	From their results, we can see that the appearance flow model is more effective, since the brightness changes are often non-linear in the endoscopic scene. Nevertheless, our MonoPCC still exceeds the AF-SfMLearner by reducing Sq Rel by 9.11\% on SCARED, which implies that compared to the learning-based appearance flow model, our cycle-form warping can guarantee the brightness consistency in a more effective and reliable way.

	Fig.~\ref{fig:scared_simcol} presents a qualitative comparison on the two datasets.
	Error maps are acquired by mapping pixel-wise Abs Rel value to different colors, and the red means large relative error and the blue indicates small error. 
	On SCARED, the color-coded error map of MonoPCC is clearly bluer than those of other SOTAs, especially for the regions near the boundaries or specular reflections, as indicated by the red dotted boxes in Fig.~\ref{fig:scared_simcol}. Specifically, the failure of other methods in these cases can be attributed to neglect or misguidance on brightness inconsistency, leading DepthNet to learn inappropriate depth prior knowledge. 
	On SimCol3D, MonoPCC provides lower relative depth errors at both the proximal and distal parts of the colon, which demonstrates the validity of MonoPCC for low-textured regions, e.g., colon in the digestive tract.

	\begin{table*}
		\small
		\centering
		\caption{The rows except the first one are the comparison results of the five variants and the complete MonoPCC, which are all cycle-constrained. The first row is the backbone MonoViT using the regular non-cycle constraint.}
		{\begin{tabular}{cccccccc}
				\toprule
				Cycle & STM & EMA & $\mathcal{L}_{pcp}$ & Abs Rel $\downarrow$ & Sq Rel $\downarrow$ & RMSE $\downarrow$  & RMSE log $\downarrow$  \\
				\midrule
				$\Circle$ & \XSolidBrush	&  \XSolidBrush   &  \XSolidBrush   & 0.106$\pm$0.051   & 1.533$\pm$1.383 & 9.737$\pm$5.221 & 0.141$\pm$0.060 
				\\
				$\CIRCLE$ & \XSolidBrush	&  \XSolidBrush   &  \XSolidBrush   & 0.110$\pm$0.047   & 1.584$\pm$1.315 & 10.003$\pm$4.930 & 0.145$\pm$0.056 
				\\
				$\CIRCLE$ & \XSolidBrush	&  \Checkmark   &  \Checkmark   & 0.108$\pm$0.048   & 1.596$\pm$1.329 & 10.047$\pm$5.335 & 0.144$\pm$0.060 
				\\
				$\CIRCLE$ &	\Checkmark &  \XSolidBrush   &   \XSolidBrush  & 0.099$\pm$0.043 & 1.286$\pm$1.091 & 9.007$\pm$4.697 & 0.131$\pm$0.052 
				\\
				$\CIRCLE$ &	\Checkmark &  \XSolidBrush   &   \Checkmark  & 0.091$\pm$0.031   & 1.040$\pm$0.700 & 8.239$\pm$3.664 & 0.120$\pm$0.041 
				\\
				$\CIRCLE$ &	\Checkmark &  \Checkmark &   \XSolidBrush  & 0.091$\pm$0.036 & 1.081$\pm$0.800 & 8.355$\pm$3.905 & 0.121$\pm$0.045 
				\\
				$\CIRCLE$ &	\Checkmark &  \Checkmark &  \Checkmark & \textbf{0.085$\pm$0.034}   & \textbf{0.953$\pm$0.712}  & \textbf{7.734$\pm$3.537} & \textbf{0.114$\pm$0.044}   \\
				\bottomrule
		\end{tabular}}
		\vspace{-0.15cm}
		\label{tab:ablation}
	\end{table*}

	\begin{figure*}[h]
		\centering
		\includegraphics[width=0.9\linewidth]{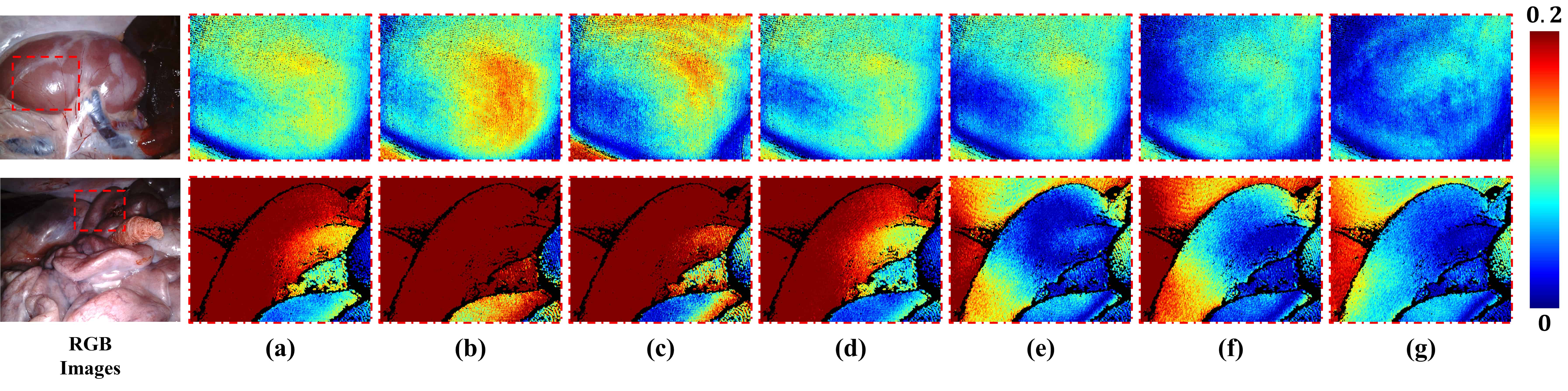}
		\vspace{-0.3cm}
		\caption{The Abs Rel error maps of seven ablation variants, including effectiveness of three components, with close-up details highlighted. (a)-(g) correspond to the $1^{st}$ to the $7^{th}$ rows in Table \ref{tab:ablation}. The regions of interest (ROIs) are outlined with red dashed lines, and the Opencv Jet Colormap is used for visualization. }
		\vspace{-0.15cm}
		\label{fig:ablation}
	\end{figure*}
	
	\vspace{-0.1cm}
	\subsubsection{Generalization on SERV-CT and Hamlyn}
	
	We also evaluate the generalization of all methods, i.e., using the model trained on SCARED to directly test on SERV-CT and Hamlyn. The comparison results are listed in Table \ref{tab:generalization}. 
	
	As can be seen, all methods experience a varying degree of degradation on both datasets. However, on SERV-CT (left part of Table~\ref{tab:generalization}), MonoPCC is still the best, and the only one maintaining the Abs Rel less than $0.1$ and $\delta$ greater than $0.9$. Also, MonoPCC significantly surpasses the second-best AF-SfMLearner by a relative decrease of 19.02\% and 11.45\% in terms of the Sq Rel and RMSE log ($p < 0.005$), respectively. 
	
	Compared to the results on SERV-CT, the performance degradation on Hamlyn is somewhat not that severe. We believe that this is because the captured scenes of Hamlyn are relatively homogeneous, without complex and irregular surfaces as SERV-CT. Compared to the second-best Endo-SfMLearner, MonoPCC achieves the lowest error in terms of the first four metrics, and the highest percentage of the inliers in terms of $\delta$ ($p < 0.001$ for all metrics).

	Fig.~\ref{fig:ct_hamlyn} exhibits four visual results from SERV-CT and Hamlyn, respectively. 
	As can be seen, compared to the other methods, MonoPCC produces less red error maps as indicated by the dotted boxes in Fig.~\ref{fig:ct_hamlyn}. 
	The two cases from Hamlyn contain motion blur during \textit{in vivo} porcine procedure (the $3^{rd}$ row), and brightness fluctuations in the over-smoothed areas of silicon heart (the $4^{th}$ row). 
	The color-coded error maps on the two challenging cases verify that MonoPCC displays almost deep blue error maps in the highlighted regions, and has better generalization on the non-rigid and reflective tissues.

	\subsection{Ablation Study}
	\label{SEC5.2}
	
	We conduct ablation studies via 5-fold cross-validation on SCARED to verify the effectiveness of each component in MonoPCC and to compare MonoPCC with other related techniques against brightness fluctuations.

	\begin{table*}
		\centering
		\caption{Comparison results of different techniques for addressing the brightness fluctuations in self-supervised learning. The last row is the technique used in MonoPCC.}
			{\begin{tabular}{ccccc}
					\toprule
					Schemes  & Abs Rel $\downarrow$ & Sq Rel $\downarrow$ & RMSE $\downarrow$  & RMSE log $\downarrow$  \\
					\midrule
					Baseline     & 0.106$\pm$0.051   & 1.533$\pm$1.383  & 9.737$\pm$5.221 & 0.141$\pm$0.060     \\
					ABT          & 0.102$\pm$0.046   & 1.425$\pm$1.214  & 9.366$\pm$4.784 & 0.137$\pm$0.056     \\
					AFM          & 0.098$\pm$0.039   & 1.251$\pm$0.964  & 8.897$\pm$4.203 & 0.130$\pm$0.049     \\
					STM+EMA & \textbf{0.091$\pm$0.036}   & \textbf{1.081$\pm$0.800}  & \textbf{8.355$\pm$3.905} & \textbf{0.121$\pm$0.045}     \\
					\bottomrule
			\end{tabular}}
			\vspace{-0.15cm}
			
			\label{tab:ablation_module}
		\end{table*}

		\subsubsection{Effectiveness of Three Components} 
		
		We develop five variants of MonoPCC by disabling the three components, EMA and/or $\mathcal{L}_{pcp}$, and STM. Note that, the warping should be cycle-form to utilize the components, and if we disable EMA, only $\theta$ in the backward warping path will be updated, and $\theta_{EMA}$ in the forward warping remains unchanged. Table \ref{tab:ablation} and Fig.~\ref{fig:ablation} give the comparison results between the variants and complete MonoPCC. We also include the backbone MonoViT in the first row of Table \ref{tab:ablation} as the baseline, which adopts the non-cycle warping.

		From the comparison results in Table \ref{tab:ablation} and visualization cases in Fig.~\ref{fig:ablation}, three key observations can be made:
		
		(1) By comparing the first two rows of Table \ref{tab:ablation} and Fig.~\ref{fig:ablation}~(a)-(b), the inferior performance of the variant using none of the components indicates that the direct usage of cycle warping is not enough due to the image blurring and unstable gradient propagation mentioned in Sec.~\ref{SEC3.2}.

		(2) As can be seen from the second to fourth rows of Table \ref{tab:ablation} and Fig.~\ref{fig:ablation}~(b)-(d), using EMA and $\mathcal{L}_{pcp}$ solely brings no improvements, which still creates much unsatisfactory estimation in the red dotted boxes of Fig.~\ref{fig:ablation}~(c).
		However, adding STM can significantly reduce Sq Rel and RMSE by 16.11\% and 7.50\%, respectively ($p < 0.003$). Thus, STM is the irreplaceable component for successfully training using the cycle-form warping.

		(3) Comparison between the last four rows of Table \ref{tab:ablation} and Fig.~\ref{fig:ablation}~(d)-(g) reveals the effectiveness of EMA and $\mathcal{L}_{pcp}$. Specifically, with STM enabled, both EMA and $\mathcal{L}_{pcp}$ bring remarkable reduction of Abs Rel by 8.08\% ($p = 0.012$, $p = 0.022$). Meanwhile, EMA and $\mathcal{L}_{pcp}$ are not mutually excluded, and the combination reduces Abs Rel by 14.14\% ($p = 0.001$). 
		As illustrated in Fig.~\ref{fig:ablation}~(g),  there exist more blue-tuned pixels in the red dotted regions of interest (i.e., either light or dark areas).

		\subsubsection{MonoPCC vs. Other Techniques against Brightness Fluctuations} 
		\label{SEC5.2.2}
		
		Several techniques have been developed to address inconsistent brightness in endoscopic images, e.g., affine brightness transformation (ABT) in Endo-SfMLearner, and appearance flow module (AFM) in AF-SfMLearner. For comparison, we train three variants of the backbone MonoViT, using ABT, AFM, and STM+EMA, respectively, to verify the robustness of different enabling techniques to the brightness inconsistency.
		
\begin{figure}[]
			\centering
			\includegraphics[width=1.0\linewidth]{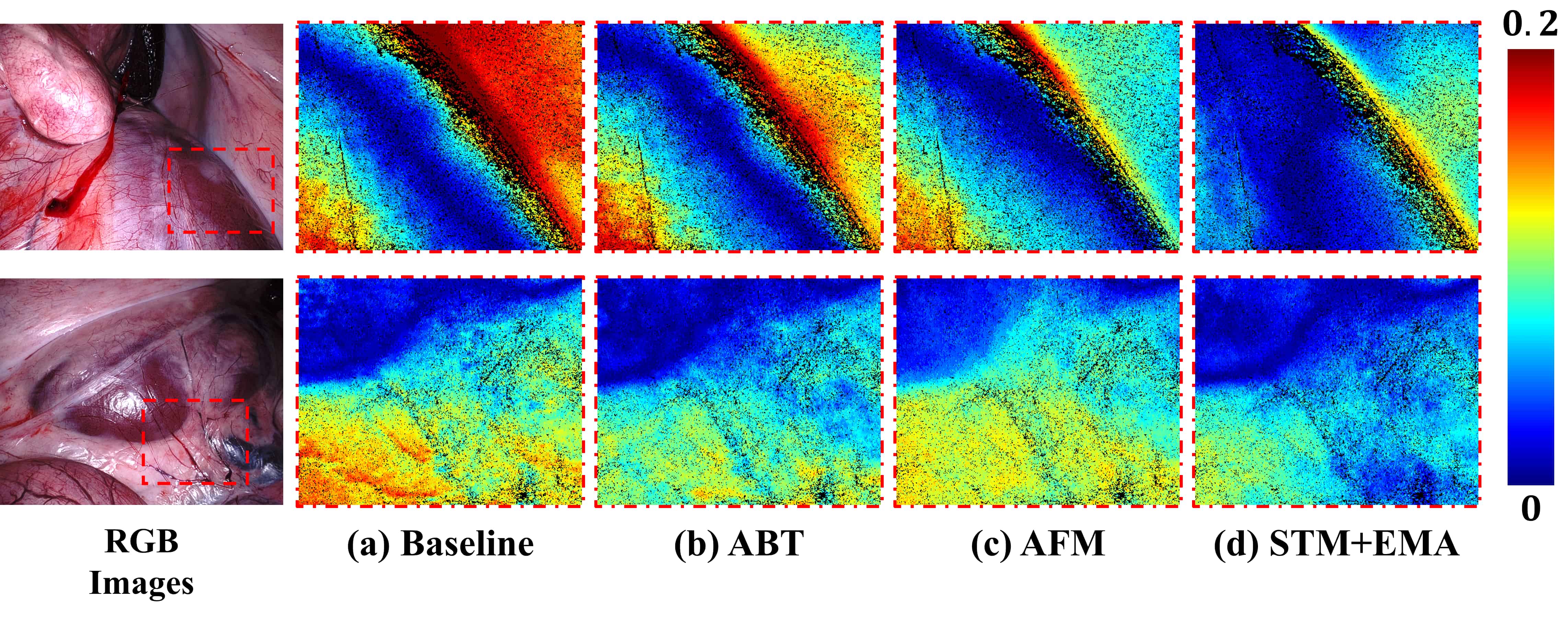}
				\vspace{-0.3cm}
			\caption{The Abs Rel error maps of MonoPCC and other similar modules against photometric inconsistency, with close-up details highlighted. (a)-(d) correspond to the $1^{st}$ to the $4^{th}$ rows in Table \ref{tab:ablation_module}. The regions of interest (ROIs) are outlined with red dashed lines, and the Opencv Jet Colormap is used for visualization. }
				\vspace{-0.3cm}
			\label{fig:ablation_module}
		\end{figure}

		Table \ref{tab:ablation_module} lists the comparison results as well as the backbone MonoViT using none of these techniques. As can be seen, the variant using STM+EMA strategy exceeds that using AFM only with a reduction of Abs Rel and Sq Rel by 7.14\% and 13.59\%, respectively ($p = 0.019$, $p = 0.014$). 
		Fig.~\ref{fig:ablation_module} displays two visualization examples in challenging cases, e.g., the boundary and flat area of tissues. As shown in Fig.~\ref{fig:ablation_module}, all techniques for addressing the brightness fluctuation can help to gain varying degrees of improvements. Furthermore, we observe that our strategy performs better than both ABT and AFM. 
		It is worth noting that compared to AFM, our method requires no extra model to learn, exhibiting greater usability and scalability.

		\subsection{Robustness to Severe Brightness Inconsistency}
		\label{SEC5.3}
		
\begin{figure}[t]
			\centering
			\includegraphics[width=0.9\linewidth]{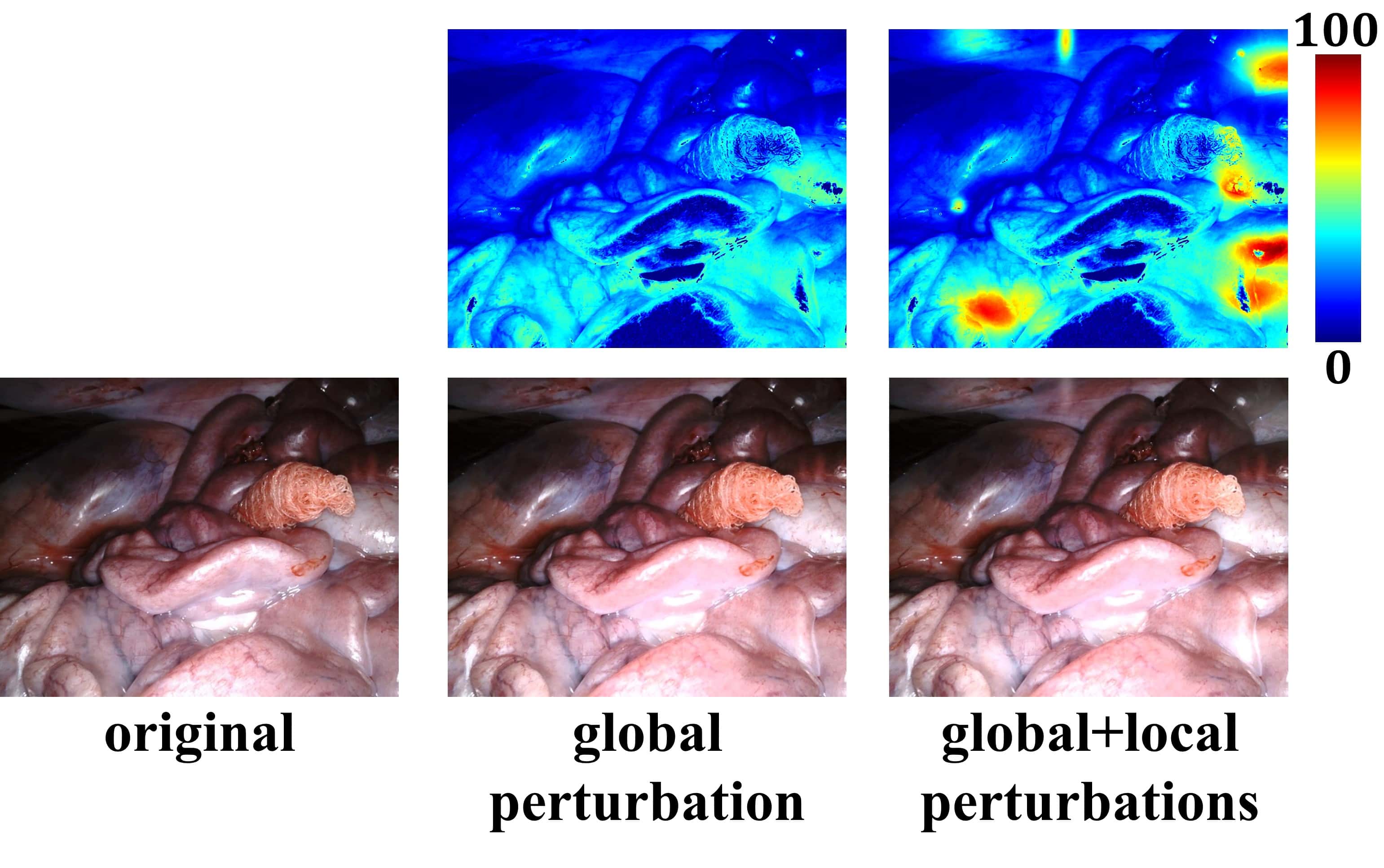}
			\caption{An example of created brightness perturbation. From left to right is the original image, globally perturbated ($k=1.2$), and both globally and locally (bright spots) perturbated. The color-coded maps above them describe the subtractive difference between the perturbated image and its original one. }
			\vspace{-0.2cm}
			\label{fig:controlled_brightness}
		\end{figure}

		Besides the current inconsistent brightness carried by the used dataset, we want to test the limit of MonoPCC under more severe brightness fluctuations. To this end, we create two copies of SCARED, and add global brightness perturbation to every adjacent frames of the first copy, and both global and local perturbations to the second. Specifically, the global perturbation is defined as the linear transformation on the image's brightness channel (HSV color model), that is, $v^{scale} = k*v, k\in [0.8, 0.9] \cup [1.1, 1.2]$. The local perturbation is the randomly placed Gaussian bright (or dark) spots. Fig.~\ref{fig:controlled_brightness} presents visual examples from the original SCARED and our created two copies with more severe brightness inconsistency.

		\begin{figure}[t]
			\centering
			\includegraphics[width=1.0\linewidth]{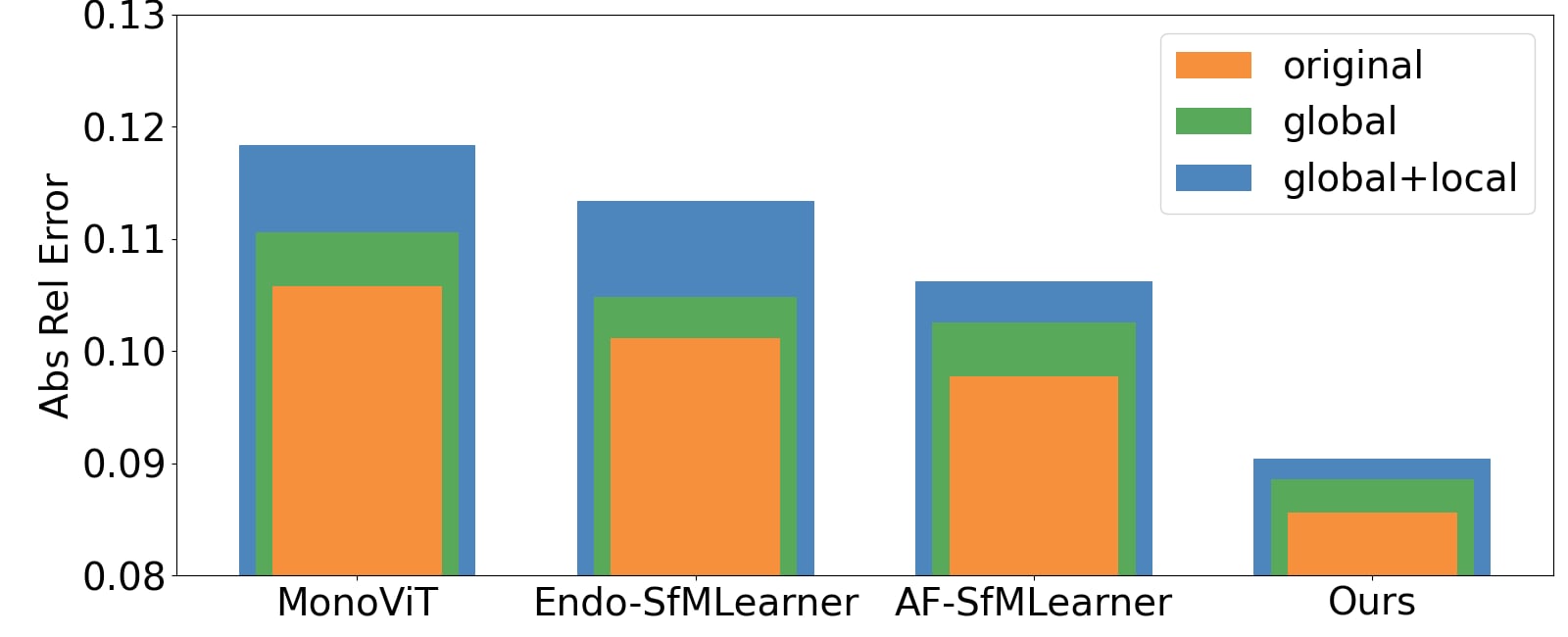}
			\caption{The Abs Rel errors of different methods trained on the two brightness-perturbed copies of SCARED and the original SCARED.
			}
			\label{fig:error_noise}
		\end{figure}
		
		Fig.~\ref{fig:error_noise} illustrates Abs Rel errors of different methods on the two brightness-perturbed copies of and the original SCARED. As mentioned before, Endo-SfMLearner and AF-SfMLearner all tried to address the brightness inconsistency. As can be seen from the comparison results, Endo-SfMLearner shows a robustness to the global brightness perturbation thanks to its using brightness linear transformation, but degrades significantly when facing the local perturbation. AF-SfMLearner can address the local brightness perturbation more or less using a learned appearance flow model, but still presents a non-trivial increase of Abs Rel error. In comparison, MonoPCC achieves the lowest Abs Rel error on the three datasets, and even produces lower errors on the copy containing the most severe brightness inconsistency (global+local) compared to the other methods' results on the original SCARED.

		\begin{table*}[t]
			\centering
			\caption{Quantitative comparison results on KITTI. The best results are marked in bold and the second best ones are underlined. }
			\begin{tabular}{lccccccc}
				\toprule
				& Year & Abs Rel $\downarrow$ & Sq Rel $\downarrow$ & RMSE $\downarrow$  & RMSE log $\downarrow$ & $\delta$ $\uparrow$    \\
				\midrule
				Monodepth2     & 2019 & 0.115 & 0.903 & 4.863 & 0.193 & 0.877   \\
				FeatDepth     & 2020 & 0.104 & 0.729 & 4.481 & 0.179 & 0.893    \\
				HR-Depth        & 2021 & 0.109 & 0.792 & 4.632 & 0.185 & 0.884    \\
				DIFFNet       & 2021 & 0.102 & 0.749 & 4.445 & 0.179 & 0.897     \\
				Endo-SfMLearner  & 2021 & 0.100 & 0.722 & 4.457 & 0.178 & \underline{0.898}          \\
				AF-SfMLearner  & 2022 & 0.113 & 1.110 & 4.949 & 0.189 & 0.883            \\
				MonoViT      & 2022 & \underline{0.099} & \underline{0.708} & \underline{4.372} & \underline{0.175} & \textbf{0.900}    \\
				Lite-Mono     & 2023 & 0.101 & 0.729 & 4.454 & 0.178 & 0.897    \\
				MonoPCC(Ours)   &  & \textbf{0.098} & \textbf{0.677} & \textbf{4.318} & \textbf{0.173} & \textbf{0.900}  \\
				\bottomrule
			\end{tabular}\\
			\label{tab:kitti}
		\end{table*}

		\subsection{Results on KITTI Benchmark}
		\label{SEC5.4}
		
		Besides the demonstration on the medical scenarios, i.e., endoscopic images, we in this subsection verify the competitiveness of MonoPCC on the popular natural-scence benchmark, i.e., KITTI \citep{geiger2012we}. KITTI is an outdoor dataset, which roughly obeys brightness consistency assumption due to constant illumination condition, i.e., the sunshine. However, some factors like specular reflection on the car body also create some negative impacts on the ideal photometric supervision. 
		
		Table \ref{tab:kitti} provides the comparison results with the previously mentioned SOTAs. Most of them have reported their evaluation results on KITTI, so we directly copy their results. 
		For the two methods, i.e., Endo-SfMLearner and AF-SfMLearner, which have no official results on KITTI, we use their released codes and default settings to train two models on KITTI.
		
		As indicated in Table \ref{tab:kitti}, MonoPCC and MonoViT are the only two methods whose Abs Rel values are less than $0.1$. Moreover, MonoPCC outperforms MonoViT in terms of all metrics, especially with a relative decrease of 4.38\% in terms of Sq Rel error ($p < 0.001$). 
		
		The two endoscopy-tailored methods, i.e., Endo-SfMLearner and AF-SfMLearner, are the second or the third best on the four endoscopic datasets, but unexpectedly slip several places on KITTI. We believe the potential reason is that the two methods are overly focusing on the brightness adjustment, and try to change the image appearance no matter the brightness is inconsistent or not. Since KITTI has no that significant brightness fluctuations, such forcible brightness adjustment could induce errors more or less. In comparison, our MonoPCC is essentially making the brightness patterns inherit across the source and target images, and thus the strength of brightness adjustment can be naturally compatible with the true situation of brightness fluctuations. Therefore, MonoPCC shows the consistent competitiveness in both medical and natural scenes.

		\begin{table}[t]
	\centering
	\caption{Quantitative comparison results (Absolute Trajectory Error) of pose estimation on two trajectories of SCARED. The best results are marked in bold and the second-best underlined.}
	\resizebox{\columnwidth}{!}
	{\begin{tabular}{lccc}
			\toprule
			& Trajectory.1 $\downarrow$           & Trajectory.2 $\downarrow$           & Average ATE $\downarrow$          \\
			\midrule
			Monodepth2      & 0.0748          & 0.0536          & 0.0606           \\
			FeatDepth       & 0.0767          & 0.0468          & 0.0567           \\
			HR-Depth        & 0.0747          & 0.0609          & 0.0654           \\
			DIFFNet         & 0.0760          & 0.0571          & 0.0633           \\
			Endo-SfMLearner & 0.0754          & 0.0518          & 0.0596           \\
			AF-SfMLearner   & \underline{0.0737}  & \underline{0.0464}  & \underline{0.0554}   \\
			MonoViT         & 0.0760          & 0.0523          & 0.0601           \\
			Lite-Mono       & 0.0758          & 0.0477          & 0.0570           \\
			MonoPCC(Ours)   & \textbf{0.0730} & \textbf{0.0459} & \textbf{0.0548} \\
			\bottomrule
	\end{tabular}} \\
\vspace{-0.1cm}
	\label{tab:pose}
\end{table}

		\begin{figure*}[t]
			\centering
			\includegraphics[width=1.0\linewidth]{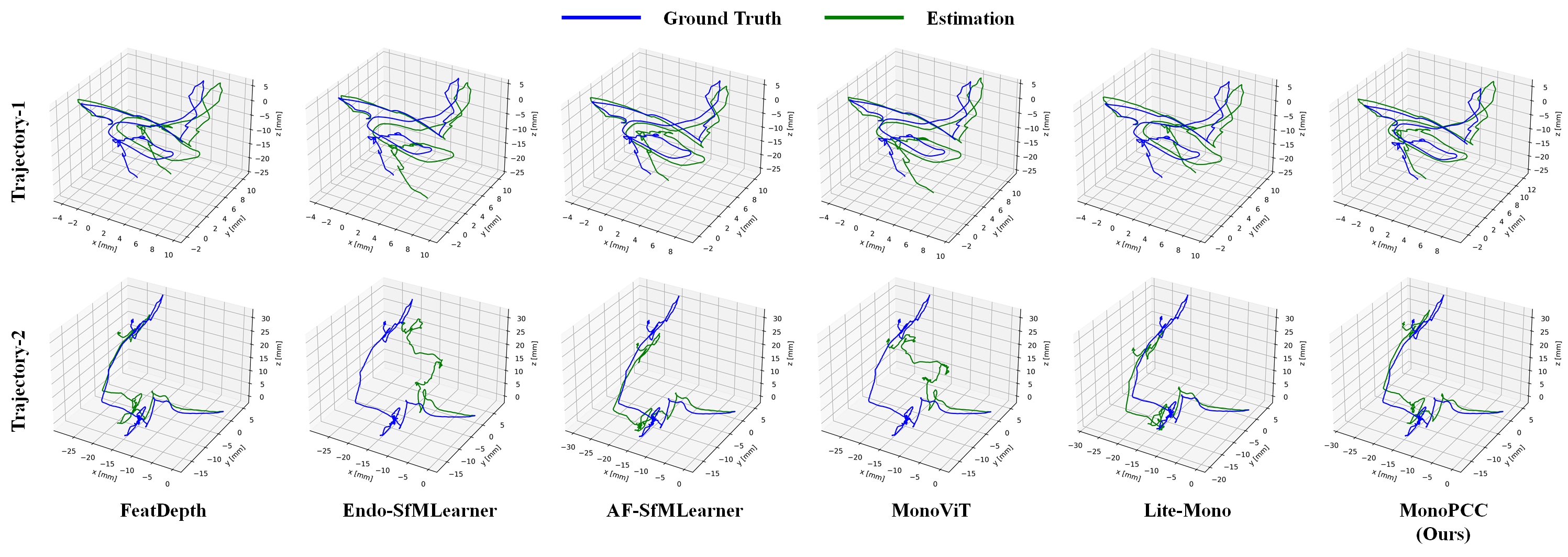}
				\vspace{-0.5cm}
			\caption{The qualitative pose estimation comparison based on two SCARED trajectories.
			}
				\vspace{-0.3cm}
			\label{fig:pose}
		\end{figure*}

		\subsection{Performance of Pose Estimation}
		\label{SEC5.5}
		
		In this subsection, we also compare the performance of ego-motion estimation between MonoPCC and other SOTA methods.
		
		Table \ref{tab:pose} presents the quantitative comparison of pose estimation on the two trajectories. While most methods, including MonoPCC, share the same PoseNet architecture, the performance of pose estimation varies significantly across methods. Notably, MonoPCC outperforms other SOTAs in pose estimation for both trajectories. Moreover, our PoseNet achieves an 8.82\% relative reduction in ATE compared to the baseline MonoViT ($p < 0.001$), a 1.08\% relative reduction compared to AF-SfMLearner ($p = 0.095$).
		
		Fig.~\ref{fig:pose} shows the predicted trajectories of our method (MonoPCC) and five other SOTAs, for two trajectories of SCARED in 3D space. The trajectories are aligned at the starting point, and the positions of the endpoints indicate how well the trajectories match the ground truth. As can be seen, the endpoints of the MonoPCC trajectory (the last column) are closer to the ground truth (the blue curve), demonstrating the superiority of our approach in pose estimation.

		\begin{figure*}[t]
			\centering
			\includegraphics[width=0.6\linewidth]{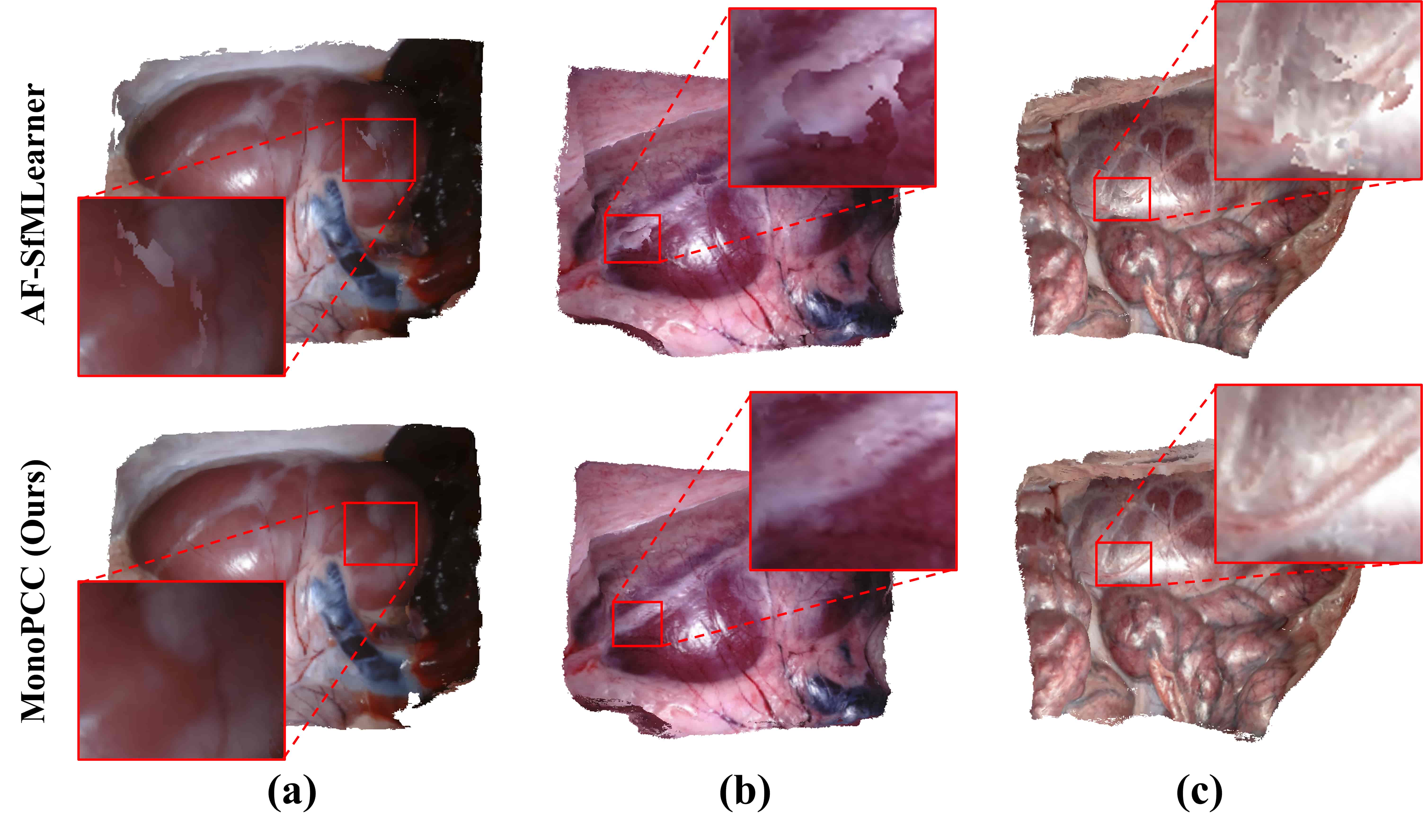}
			\caption{Qualitative comparison results on the 3D scene reconstruction based on the estimated depth maps of two methods. The three sequences are selected from SCARED.
			}
			\label{fig:reconstruction}
		\end{figure*}

		\subsection{Performance of Scene Reconstruction}
		\label{SEC5.6}
		
		Using the known camera intrinsics and the estimated depth maps from continuous video frames, we aim to recover 3D surface models and merge them into a larger scene with richer spatial information. Specifically, we estimate the depth maps of three sequences from the SCARED dataset using the top-performing methods: MonoPCC and AF-SfMLearner. Following the setting of Endo-Depth-and-Motion \citep{recasens2021endo}, we apply photometric tracking and volumetric reconstruction to fuse the depth maps and generate 3D mesh models. 
		
		Fig.~\ref{fig:reconstruction} presents three examples of the reconstructed scenes, along with close-up views of the details. In general, both depth estimation methods provide relatively good results for 3D reconstruction. However, by comparing the highlighted regions in Fig.~\ref{fig:reconstruction}, we observe that the surface meshes generated by AF-SfMLearner exhibit noticeable artifacts. These artifacts are likely due to depth outliers in the predicted maps. In contrast, our method (MonoPCC) produces smoother and more accurate 3D reconstructions, demonstrating its potential for high-quality 3D scene reconstruction.

		\section{Conclusion}
		Self-supervised monocular depth estimation is challenging for endoscopic scenes due to the severe negative impact of brightness fluctuations on the photometric constraint. 
		In this paper, we propose a cycle-form warping to naturally overcome the brightness inconsistency of endoscopic images, and develop a MonoPCC for robust monocular depth estimation by using a re-designed photometric-invariant cycle constraint.
		To make the cycle-form warping effective in the photometric constraint, MonoPCC is equipped with two enabling techniques, i.e., structure transplant module (STM) and exponential moving average (EMA) strategy.
		STM alleviates image detail degradation to validate the backward warping, which uses the result of forward warping as input.
		EMA bridges the learning of network weights in the forward and backward warping, and stabilizes the intermediate warped image to ensure an effective convergence.
		The comprehensive and extensive comparisons with $8$ state-of-the-arts on five public datasets, i.e., SCARED, SimCol3D, SERV-CT, Hamlyn, and KITTI, demonstrate that MonoPCC achieves a superior performance by decreasing the absolute relative error by 7.27\%, 9.38\%, 9.90\% and 3.17\% on four endoscopic datasets, respectively, and shows the competitiveness even for the natural scenario.
		Additionally, two ablation studies are conducted to confirm the effectiveness of three developed modules and the advancement of MonoPCC over other similar techniques against brightness fluctuations.

		\textbf{Limitations. }The current pipeline relies on a single frame to infer the depth map. Since each prediction is made independently, the model lacks perception of temporal consistency, meaning that the depth values at the same location may vary over time. This temporal inconsistency can lead to artifacts, such as overlapping tissue surfaces, as shown in the 3D reconstruction visualization in Fig.~\ref{fig:reconstruction}. Furthermore, our method is primarily designed for static endoscopic scenes. In dynamic scenarios involving tissue deformation, MonoPCC may not perform effectively. This is because the depth values at corresponding positions between source-target paired images can change locally, making it difficult to establish the cycle warping path consistently.
		
		\textbf{Potential Future Application. }In this paper, we have demonstrated the effectiveness of the PCC strategy for self-supervised monocular depth estimation in endoscopic images. We believe that our framework can be seamlessly integrated into other related tasks, such as stereo matching \citep{shi2023bidirectional} and metric depth estimation \citep{wei2022distilled,wei2024absolute}, both of which face challenges due to brightness fluctuations. Additionally, in the field of NeRF-based scene reconstruction, several depth-prior-assisted methods \citep{wang2022neural,li2024endoself,huang2024endo} utilize estimated depth to guide model training. Therefore, our depth estimator, designed specifically for endoscopic scenes, could also enhance the performance of such downstream tasks.

		\section*{Acknowledgments}
		This work was supported in part by National Key R\&D Program of China (Grant No. 2023YFC2414900), Fundamental Research Funds for the Central Universities (2021XXJS033), Research grants from United Imaging Healthcare Inc.

		\bibliographystyle{model2-names.bst}\biboptions{authoryear}
		\bibliography{medima-template}
		
		%
		
	\end{document}